\documentclass{svproc}
\usepackage{graphicx}
\usepackage{subfigure}
\usepackage{booktabs} 

\usepackage{amsmath}
\usepackage{amsfonts}
\usepackage{mathtools}
\usepackage{dsfont}
\usepackage[square,numbers]{natbib}
\usepackage[hidelinks]{hyperref}
\usepackage{float}

\usepackage{url}

\begin{document}
\mainmatter              
\title{Achieving Goals using Reward Shaping and Curriculum Learning}
\titlerunning{Achieving Goals}  
%
\author{Mihai Anca\inst{1} \and Jonathan D. Thomas\inst{1} \and Dabal Pedamonti\inst{1} \and Mark Hansen\inst{2} \and Matthew Studley\inst{2}}
\authorrunning{Mihai Anca} 
%
\tocauthor{Mihai Anca, Jonathan D. Thomas, Dabal Pedamonti, Mark Hansen and Matthew Studley}
\institute{University of Bristol, Bristol, United Kingdom\\
\email{mihai.anca@bristol.ac.uk}
\and
University of the West of England, Bristol, United Kingdom}

\maketitle              

\begin{abstract}
    Real-time control for robotics is a popular research area in the reinforcement learning community. Through the use of techniques such as reward shaping, researchers have managed to train online agents across a multitude of domains. Despite these advances, solving goal-oriented tasks still requires complex architectural changes or hard constraints to be placed on the problem. In this article, we solve the problem of stacking multiple cubes by combining curriculum learning, reward shaping, and a high number of efficiently parallelized environments. We introduce two curriculum learning settings that allow us to separate the complex task into sequential sub-goals, hence enabling the learning of a problem that may otherwise be too difficult. We focus on discussing the challenges encountered while implementing them in a goal-conditioned environment. Finally, we extend the best configuration identified on a higher complexity environment with differently shaped objects.
\keywords{reinforcement learning, curriculum learning, reward shaping, robotics}
\end{abstract}

\section{Introduction}

Long-horizon manipulation tasks, such as object stacking, require planning over an extended period of time while taking into account various mandatory checkpoints. Creating agents that can solve this type of task using complex reward signals is a key challenge in deep reinforcement learning (deep RL). Recent advances have shown that RL applied to robotic arms can perform object stacking \cite{jeong2020self, hundt2020goodR, pmlr-v164-lee22b, li2020towards}, however, the problem is usually constrained, or the solution requires complex changes to the model's architecture. Moreover, standard RL approaches do not scale well when the number of task objects is increased.

We believe that a curriculum learning (CL) framework can help alleviate this issue by decomposing the task into sub-goals of incremental difficulty. This has been shown by \citet{narvekar2020curriculum} to be effective in applications such as robotics and games.

We leveraged the use of the simulation platform Isaac Gym \cite{makoviychuk2021isaac}, which allows thousands of environments to be asynchronously controlled by an RL agent. We found that there is a gap in the on-policy RL literature when dealing with heavily parallelized object-centric goal-oriented tasks, such as stacking. Our work, to the best of our knowledge, provides the first results in this territory.

Given that we used an online on-policy algorithm, the methodology and findings described in this paper can be transferred to other sample-efficient algorithms that train agents directly in the real world.

In summary, the paper makes the following contributions: 1) We present various CL setups and couple them with reward shaping, enabling the stacking of three objects using standard proximal policy optimization (PPO) \cite{schulman2017ppo}. To the best of our knowledge, no prior work successfully achieved this task using on-policy vanilla RL algorithms (i.e. PPO). 2) We analyse the elements that hinder learning and discuss the key components that are critical for task completion. 3) We show that in using an asynchronous setup, such as Isaac Gym {\cite{makoviychuk2021isaac}}, to learn goal-oriented tasks, the learning horizon becomes an important hyperparameter. We then consider the effects of a curriculum for adjusting this value. 4) Finally, we provide the \textit{\hyperlink{https://github.com/MihaiAnca13/AchievingGoalsCL}{codebase}} to the community to act as a baseline for future work in multiple cube-stacking tasks.

\textbf{A surprising finding:} Choosing the optimal learning horizon value for the whole task of stacking three cubes results in a better convergence time when compared to selecting the optimal value per curriculum stage. Allowing 50 timesteps for placing each object, the agent trains faster and achieves a higher accuracy if trained from the start with a horizon length of 150 (assuming three objects in the scene). This finding is discussed in Section \ref{s:horizon}.

\section{Background and Related Work}
The ``Pick and Place'' task was popularised as part of a reinforcement learning (RL) baseline with the release of OpenAI robotics gym pack \cite{brockman2016openai} in 2016. Both online and offline RL methods have been shown to work on the dense reward version of this environment. For example, in \citet{jeong2020self}, the authors used a model-based vision approach and self-supervision to pick and place one cube on top of another. 

However, the literature to date offers no simple solution to the problem of stacking at least three objects, which represents a key example of a long-horizon task. In \citet{hundt2020goodR}, the authors discretised the action space and managed to learn an RL policy that successfully stacks multiple objects. This method relies on prior knowledge and is only applicable to simple-to-grasp objects, such as cubes. 
The recent work of \citet{pmlr-v164-lee22b} showed a potential solution when stacking differently shaped objects, however, the problem solved is heavily constrained and requires multiple training sessions.

This problem becomes increasingly difficult when posed in a sparse reward environment. \citet{li2020towards} overcame it by using graph neural networks (GNN) coupled with attention layers, successfully stacking up to six cubes in simulation. However, even though the number of trainable parameters is similar to a fully connected network, the new architecture requires considerably more VRAM when computing the message-parsing step of the GNN. Additionally, it failed to explain the underlying issue of why vanilla RL approaches are incapable of learning this task.

Another popular example of stacking multiple cubes using sparse rewards is that presented in \citet{nair2018overcoming}. The authors made clever use of the critic network in an actor-critic architecture to label the importance of demonstrations. Coupled together with Hindsight Experience Replay (HER) \cite{andrychowicz2017hindsight}, the trained agent learned how to stack up to six cubes, although it required the production and presentation of 100 demonstrations.

The question is, therefore, whether RL is capable of solving this stacking problem with the use of complex reward shaping and CL, as opposed to highly constrained and/or architecturally intricate scenarios or with the help of demonstrations. Examples of a complex problem solved using reward shaping and RL include \citet{bellemare2020autonomous} and \citet{hu2020learning}. The advantage that this architectural simplicity brings is a reduction of pipeline complexity for reproducibility and transferability, which can aid with safety constraints that can be added to the system later on. 


\subsection{Reinforcement Learning}

Throughout the paper, we will consider the goal-conditioned RL formalism \cite{kaelbling1993learning} with a fully observable environment. 
The goals are individually sampled from a uniform distribution at the beginning of each episode.
We utilise proximal policy optimization (PPO) \cite{schulman2017ppo} with the generalized advantage estimator (GAE) \cite{schulman2015gae} to find the optimal policy.
Our environment is built as an infinite-horizon setting with early stopping. We
chose this setup because it significantly speeds up learning due to focusing the exploration around the starting position of the agent.

\subsection{Curriculum Learning}

Given a collection of tasks, each characterized by a reward function dependent on a target goal, a simple solution would involve attempting to learn all tasks simultaneously. This, unfortunately, is inefficient due to ignoring the dependencies between the tasks. Moreover, this solution assumes the agent is capable of learning any of these tasks in isolation, irrespective of the agent's current capabilities. However, if the fraction of learnable tasks is small, this becomes computationally inefficient.

A better alternative is presented through the Curriculum Learning (CL) paradigm. A curriculum can be formed to narrow the distribution of tasks  by training the agent only on tasks that it is capable of learning. Generally, it is unknown in advance which tasks can be learned given the current agent's skills. However, there are environments, such as stacking of items, where prior knowledge can be utilised to select the right order for the tasks in which they are presented to the agent (i.e. gradually increase the number of items in the scene).  

In other words, CL is a training procedure that increases performance and/or training speed on a high-complexity task by decomposing it into a sequence of lower-difficulty sub-goals. Due to increased sample efficiency, the agent amasses knowledge faster and is able to generalize towards the complex task, reducing the exploration required if it were to train uninitialized. 

\begin{figure*}[!t]
    \centering
    \includegraphics[width=0.9\textwidth]{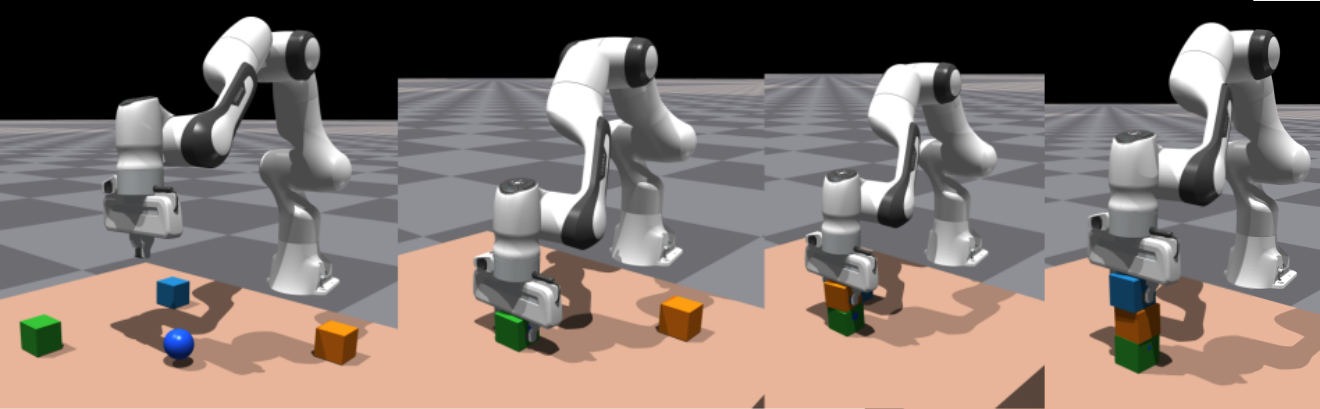}
    \caption{We present our setup using a Franka Panda arm with 3 cubes in a simulation, created in Isaac Gym \cite{makoviychuk2021isaac}. The image shows the intermediary goals achieved by a trained agent when stacking 3 objects.}
    \label{fig:env_stages}
\end{figure*}

The first application of CL in robotics dates back to 1994 in \citet{sanger1994firstCL}, where a parameter was used to change the difficulty of the environment in which the agent operates. Other influential work includes \citet{schmidhuber2013powerplay}, in which increasingly difficult sub-problems were identified and trained on. \citet{florensa17a} presented a novel and practical approach for sampling a start-state distribution that varies over the course of training, leading to an automatic curriculum of start-state distributions. Curriculum-based approaches with manually designed schedules have also been explored in supervised learning \cite{graves2017automated}. More recently, \citet{forestier2021intrinsicCL} demonstrated how CL can be combined with intrinsic motivation to overcome certain robotics tasks. 

For a more comprehensive overview of curriculum learning, please refer to  \citet{narvekar2020curriculum} and \citet{wang2021survey}.

\subsection{Reward Shaping}

Goal-oriented tasks are naturally described using sparse reward functions where the agent receives a positive reward only when achieving the task, and zero otherwise. Unfortunately, this makes the learning of such tasks difficult, due to the amount of guided exploration required to reach the goal even once. Most commonly \cite{ng1999originalShaping}, the process of reward shaping is used to overcome this issue. Reward shaping refers to modifying the original sparse reward function by incorporating domain knowledge in the form of additional reward.

\citet{amodei2016concrete} and \citet{grzes2017rewardShaping} observe that there are downsides to using a shaped reward. Without sufficient care, the changes brought to the reward function can mislead the agent towards a policy optimised for a local optimum introduced accidentally through the reward shaping. We explore these implications through an ablation study, which can be found in the Appendix.

\section{Methods}

The environment we use (Fig. \ref{fig:env_stages}) in our experiments consists of a Franka Panda (7 DOF) robotic arm, a table, and 3 boxes. The goal is to stack the objects on top of each other. The goal location and initial object positions are sampled randomly in every episode within the confines of the table. The table surface is split into a 5x5 grid, where each square represents a possible initial location for one of the cubes, thus preventing collisions from taking place during initialisation.

The initial joint positions are obtained by setting them to a hard-coded home position to which a small random perturbation (8\%) is added: $s_0 = s_{home} + 0.08 \cdot L \cdot \mathcal{U}[-1,1]$, where $L$ represents the joint ranges. A total of 9 continuous actions are used, which include joint positions and gripper control. The observations are continuous and consist of joint position and velocity, end-effector position, object states, and goal position. More information can be found in the Appendix.

We implement the environment described above in Isaac Gym \cite{makoviychuk2021isaac}, an end-to-end high-performance robotics simulation platform. Throughout our experiments, we use the RL Games \cite{rlGames} (a highly-optimized GPU RL framework) implementation of Proximal Policy Optimization algorithm \cite{schulman2017ppo}. The Appendix contains more details regarding the hyperparameters used.

\subsection{Reward function} \label{s:reward_f}

Similar to \citet{bellemare2020autonomous} and \citet{hu2020learning}, the objective of the environment is encoded using a shaped reward function: $r_t = \sum_{i \in \mathbb{S}}{\lambda_i R_i}$, where $\mathbb{S}$ represents the reward function set (see Table \ref{t:reward} for the $\lambda$ values used). We start with the standard sparse reward for our environment, which rewards the agent when all the objects $s_\text{obj}$ are placed at their associated goal positions $s_g$ as shown in Eq. \ref{eq:sparse}. We then add a dense reward as shown in Eq. \ref{eq:dense}, represented by the mean squared error between the current and target position of the objects. This error is summed across all objects.

\begin{equation} \label{eq:sparse}
    r_{\text{sparse}} = \prod_{\text{obj} \in \mathbb{O}}{ \mathds{1}\{s_{\text{obj}} \approx s_g \pm \epsilon\} }
\end{equation}

\begin{equation} \label{eq:dense}
    r_{\text{dense}} = -\sum_{\text{obj} \in \mathbb{O}}{ \| s_{g} - s_{\text{obj}} \|_2^2 }
\end{equation}%

Additionally, we use a similar dense reward to guide the agent's end effector $s_\text{ee}$ toward the objects $s_\text{obj}$ in the scene: $ r_{\text{guide}} = -\sum_{\text{obj} \in \mathbb{O}}{ \| s_{ee} - s_{\text{obj}} \|_2^2 }$.

We introduce three additional signals for keeping the agent's choice of action as safe and efficient as possible. The first reward signal punishes wasteful actions $a$ and joint positions $\theta$ that are far from the initial values: $r_{\text{action}} = -\| a\|_2^2 - \| \theta - \theta^0\|_2^2$. The second signal gives a high punishment whenever the end effector's position on the Z axis $s_Z^\text{ee}$ surpasses a threshold $\Theta$, equal to the table height: $ r_{\text{table}} = -\mathds{1}\{s_Z^\text{ee} \le \Theta \}$. This prevents the agent from touching the table, which would otherwise cause problems when attempting to transfer the policy to a real robot. Finally, the third signal rewards the agent for maintaining the end effector orientation $\omega_\text{ee}$ close to the one it was initialised with $\omega^0$: $r_\text{orientation} = - (\omega^0 * \hat{\omega_\text{ee}})$. The orientations are represented in quaternions, and $\hat{\omega_\text{ee}}$ symbolises the conjugate.

\begin{table}[h!]
\centering
\caption{Reward function set $\mathbb{S}$}
\label{t:reward}
\resizebox{0.6\textwidth}{!}{%
\begin{tabular}{lll}
\hline
\textbf{Signal}              & \textbf{$\lambda$} & \textbf{Type} \\ \hline
Goal achieved                & 150            & $r_{\text{sparse}}$        \\
One time bonus per sub-goal  & 150            & $r_{\text{sparse}}$        \\
Box to goal distance         & 5             & $r_{\text{dense}}$         \\
End-effector to box distance & 5             & $r_{\text{dense}}$         \\  
Action magnitude penalty     & 0.01          & $r_{\text{action}}$         \\
Touching table               & 5             & $r_{\text{table}}$        \\
Orientation error            & 0.1             & $r_{\text{orientation}}$         \\ \hline
\end{tabular}
}%
\end{table}

\subsection{Types of curriculum} \label{s:curriculum_types}
In our experiments, we explore two curriculum setups. The switch between stages of a curriculum takes place automatically whenever the agent has reached an accuracy that surpasses a pre-set threshold. The accuracy is calculated as a moving average, updated each time one of the environments finishes an episode. In our experiments, we found that a threshold of 90\% strikes the balance between time and optimality.

Our first curriculum setup consists of modifying the number of active objects in the scene. All three cubes are present on the table at all times, and their positions are used as part of the observation. The initial position of the cubes is selected from the 5x5 grid explained at the beginning of this section. This setup maintains an unchanged observation space throughout all stages of the curriculum.

The change, however, comes in the form of a gated reward function $r_{\text{gated}} = \sum_{i}^{|\mathbb{O}|}{ r_\text{sparse} \cdot \mathds{1}_{i<=S}}$, such that at each stage $S$, the agent is encouraged to place one more object than in the previous stage. If a correctly placed object rewards the agent, we label them as \textit{active} for the current curriculum stage. Otherwise, we call them \textit{inactive}.

In our experiments, we gather trajectories for a number of steps equal to the horizon length $H$, allowing us to maintain a fixed batch size. The value of $H$ can be tuned depending on the task, which introduces a new hyperparameter, requiring calibration. The only constraint we have is that $H$ must be smaller than the episode length.

Our second curriculum setup builds on top of the gated reward function introduced above. It adjusts the episode and horizon length, such that, at each stage, these values are updated based on the required time for solving the task with the active number of objects.

\section{Input remapping}
 
As a baseline, we start with what we call an absolute reward, meaning all rewards introduced in Section \ref{s:reward_f} are enabled. In this scenario, the episode only terminates when all present objects are positioned at their associated targets, or when the number of steps taken surpasses the episode length.

The results (Table \ref{tab:curriculum1}) show that this baseline is incapable of learning anything meaningful during training. The gated reward curriculum, introduced in Section \ref{s:curriculum_types}, represents our first proposed improvement. In this curriculum setup, as opposed to using the absolute reward, the episode ends when all \textit{active} objects in the current curriculum stage are correctly placed.

Our first iteration establishes an order for the boxes to be placed in, similar to \cite{li2020towards}. Moreover, we disable the observations associated with objects whose reward is gated, by replacing them with 0s. For example, during the first stage of the curriculum, only the red box contributes towards calculating the reward, and the observations of the other two inactive boxes are replaced with 0s before being passed to the agent.

In this iteration, the agent learns the sequence of actions that leads to the goal only with respect to the first box. When the second stage of the curriculum is reached, and the second object activates, the learned policy fails to complete the task associated with the first stage (i.e. place the red box at the bottom of the stack). This is because activating the second set of observations creates a strong shift in the input distribution to both policy and value function. Consequently, the generalization of knowledge from task 1 to task 2 is extremely dependent on the (often problematic) out-of-distribution generalization capability of the employed function approximators. This is similar to the problems encountered in the domain of continual learning \cite{lesort2020continual}.

Unfortunately, simply allowing the state of the inactive objects to contribute to the agent's observation (instead of replacing them with 0s) does not have any beneficial effect on training. We explore a solution throughout the next sections.

\subsection{Related Methods}
The problem of input remapping has been studied on a different set of environments and confirmed by \citet{tang2021permutation}. Similarly, in \citet{li2020towards}, the authors achieved high accuracy in stacking up to 6 cubes, by using a curriculum and an architecture of GNNs with attention. GNNs provide input permutation invariance, thus overcoming the problem of input remapping. Our work differs through architectural simplicity and the insights discovered that enabled the learning to take place.

Alternative solutions include the use of elastic weights consolidation \cite{kirkpatrick2017ewc}, or of an auto-encoder \cite{raffin2019decoupling}. The first method identifies the weights that are essential to previously learned tasks and reduces further changes, such that the behaviour is preserved. This is achieved by freezing the node weights that have contributed most to learning that task. The second method overcomes the remapping issue by translating the states into a latent representation, formed using a variational auto-encoder, that is invariant to the permutation of inputs. Both of these solutions require architectural changes and may be explored in more detail in a separate study.

Our hypothesis is that the lack of permutation invariance of standard neural networks is one of the reasons why vanilla RL methods were not capable of solving the problem of stacking multiple objects, particularly in a curriculum setting. Without addressing this issue first, the curriculum of sub-tasks brings almost no benefits to the training, since the trajectory of rewarding actions has to be learned from scratch whenever a new object is introduced.

\subsection{Results} \label{s:ir_results}

We identified a solution to the input remapping issue presented above that requires no architectural changes, however, all objects must be present from the start. This is overcome by crafting the reward function in a way that guides the agent to interact with all the boxes even in the first stage of the curriculum. To achieve this, every time the episode is reset, a permutation of boxes is chosen randomly, such that a different stacking order is required for achieving the goal. As a result, all input weights associated with objects are trained equally. 

Therefore, our final iteration for this first curriculum setup consists of enabling the observations for all objects, where the order for the gated reward used is permuted for each episode. An episode ends when either enough timesteps have passed, or all the \textit{active} objects in the current stage have been positioned correctly.

Lastly, to better understand the performance boost brought by the curriculum, we used a staggered reward as a baseline, similar to the one described in \cite{bellemare2020autonomous}. Here, we still make use of the gated reward function, however, the episode only ends when all three objects are correctly positioned. See Table {\ref{tab:curriculum1}} for a comparison of our methods.

The curriculum run that achieved the best results was trained for approximately 21.5 billion steps, which is equivalent to 20 hours on our workstation (see Appendix for hardware details).

\begin{table}[h!]
\centering
\caption{Results of the various iterations of the first curriculum}
\label{tab:curriculum1}
\resizebox{0.8\textwidth}{!}{%
\begin{tabular}{lllllllll}
\hline
\textbf{Name} & \textbf{Reward} & \textbf{Input} & \textbf{Ep. stop} & \textbf{Obj. order} & \textbf{Acc.} & \textbf{Ep. L.} & \textbf{Steps} & \textbf{Time} \\ \hline
Absolute      & normal  & all & all obj & fixed       & 0\% & - & 30B & 45h \\
Curriculum 1a & gated & padded  & active obj & fixed & 0\% & - & 30B & 28h \\
Curriculum 1b & gated & all & active obj & fixed & 0\% & - & 30B & 28h \\
Staggered     & gated & all & all obj & fixed   & 81\% & 90 & 30B & 31h \\
Curriculum 1c & gated & all & active obj & permutation & \textbf{90\%} & \textbf{82} & \textbf{21.5B} & \textbf{20h} \\ \hline
\end{tabular}%
}
\end{table}

\section{Horizon length} \label{s:horizon}

\begin{figure}[!tbp]
   \centering
    \includegraphics[width=.68\textwidth]{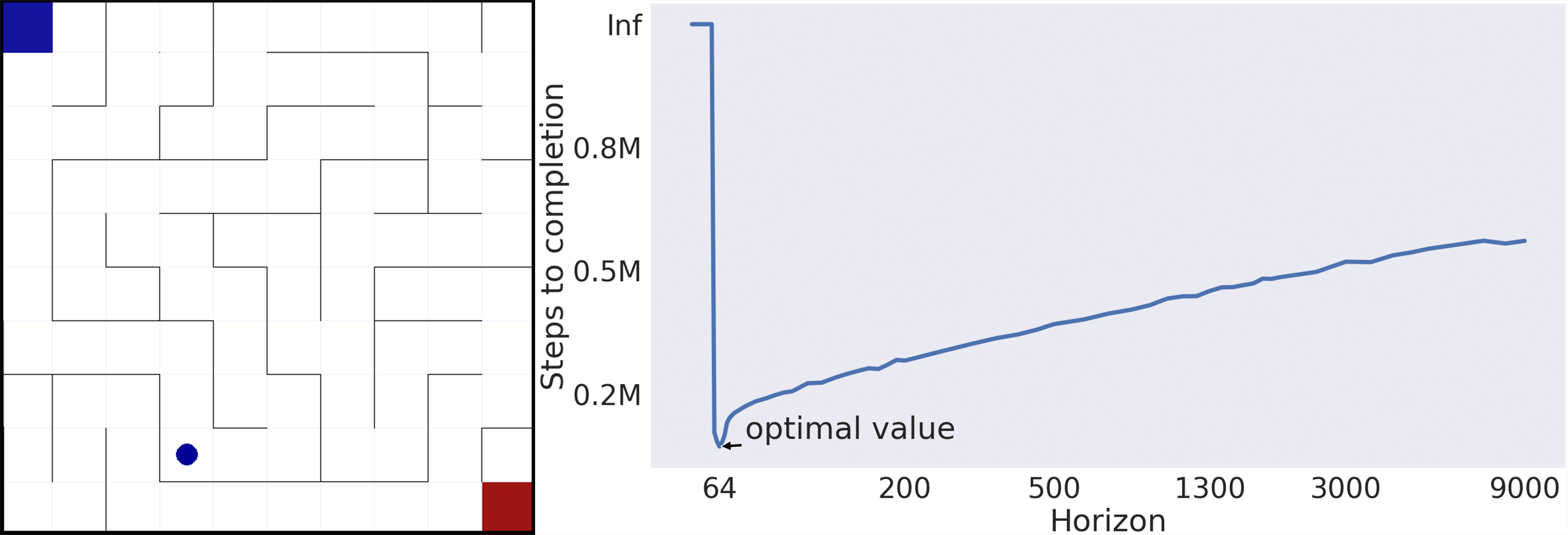}
  \caption{Left: 10x10 maze environment with discrete action space using sparse rewards provided when the agent reaches the goal (red square). Right: Training steps required to learn a policy capable of solving the maze environment over a range of values for horizon length. Providing more time in each episode seems to have diminishing returns for the overall training time.}
  \label{fig:maze-results}
\end{figure}%

PPO \cite{schulman2017ppo} is an on-policy online RL method, meaning that updates take place after each episode terminates using the gathered transitions. The algorithm was designed with parallelism in mind, where each agent collects the same amount of $H$ timesteps before each update. Further studies \cite{stooke2018accelerated} into the efficiency of parallelization considered only score-based environments, such as the Pong and Breakout games, but failed to analyse the importance of the horizon window when dealing with goal-oriented scenarios. In their bipedal locomotion environment, the agent can make incremental improvements as long as it is not in an unrecoverable state (e.g. fallen). However, we show in this section that in a goal-oriented task, the size of the learning horizon $H$ is an important hyperparameter that offers a trade-off between optimality and robustness. While this may seem obvious to the reader, there have been no published results in this area.


Our second curriculum setup consists of increasing the horizon and episode length at each stage, such that the agent has more time to explore and solve the task before being reset. Our hypothesis is that choosing a value of $H$ close to optimal for solving the task at each stage will decrease the training time.

Ideally, we want to select the optimal $H$ value for all tasks. Unfortunately, this value is hard to determine without solving the task beforehand, especially for long-horizon tasks requiring multiple checkpoints to be reached, such as object stacking. Therefore, we will next study the effect of the horizon length on convergence speed using a toy example (see Fig. \ref{fig:maze-results}).

The results show that there is an optimal value and any further increase provides diminishing returns, leading to a less-than-optimal training time. Intuitively, if $H$ drops below the required steps for solving the task, the trajectories gathered do not encapsulate the solution, hence making the problem impossible to learn. Knowing this, we empirically estimated a base horizon length $H_\text{base}$ of 50 steps per \textit{active} object. This is a good trade-off between the risk associated with having a too-small horizon length, and the sub-optimality of a too-large value. 

The transition between stages in this setup is identical to the first curriculum, presented in Section \ref{s:curriculum_types}. The horizon length is updated as follows: $H = H_\text{base} * S$, where $S$ represents the stage number.

\subsection{Reward hacking}

\begin{figure}[h!]
    \centering
    \includegraphics[width=.363\textwidth]{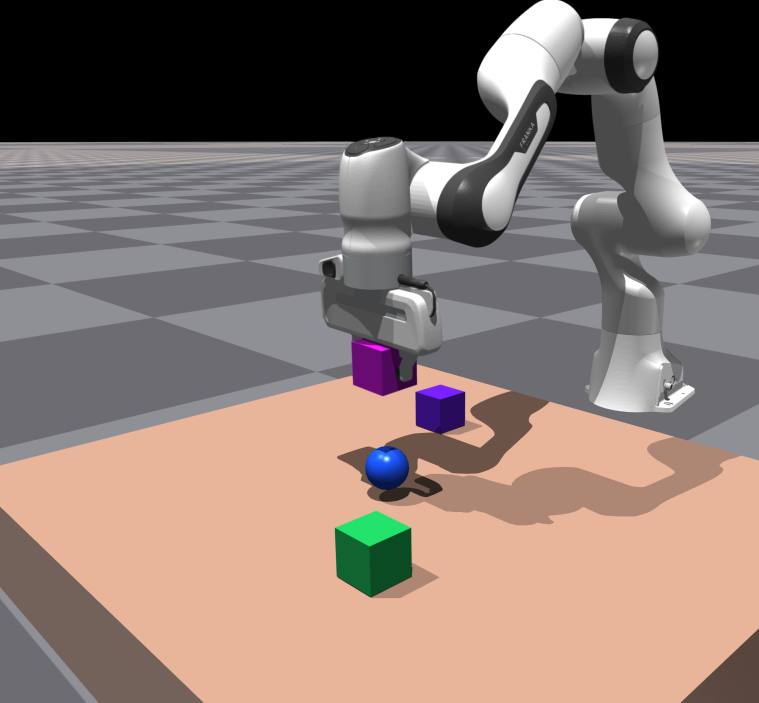}
    \includegraphics[width=.408\textwidth]{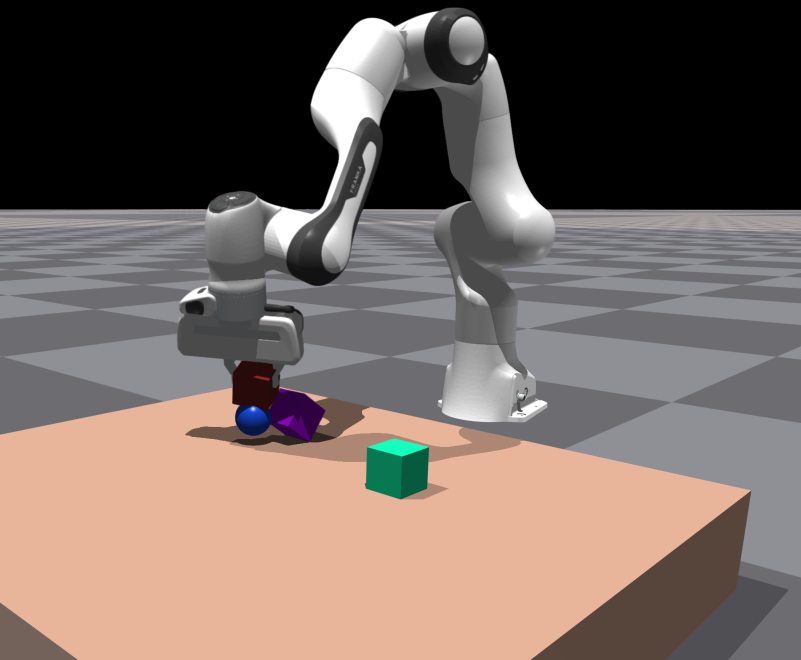}
    \caption{The first image shows an example of reward hacking where the agent is using a single cube to collect all three one-time bonus rewards. The second image shows the robotic arm pushing aside the first correctly placed box, therefore postponing the end of the episode.}
    \label{fig:reward_hacking}
\end{figure}%

Throughout our experiments concerning the second curriculum, we came across a couple of reward-hacking scenarios (see Fig. \ref{fig:reward_hacking}) directly linked to the tuning of the horizon length. This section is dedicated to explaining these instances.

In \citet{stooke2018accelerated}, each parallel agent has access to its own simulator where experiences are gathered asynchronously from the learner and resets can happen in sync with data acquisition. Isaac Gym \cite{makoviychuk2021isaac} reaches a high number of parallel environments by creating them all using the same simulator. As a consequence of maintaining a fixed batch size, the training takes place after the number of steps dictated by the horizon length has been gathered across all environments.

\begin{figure}[!tbp]
  \centering
  
  \includegraphics[width=0.72\textwidth]{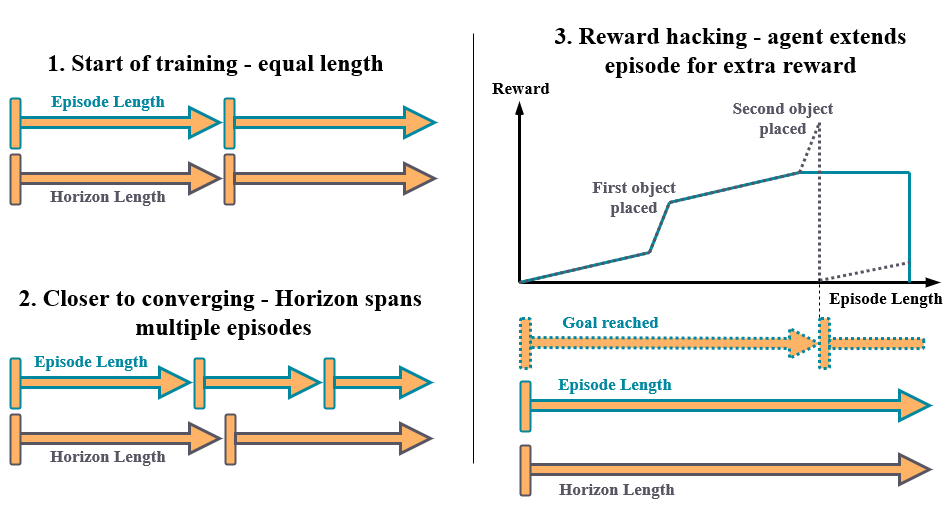}
  \caption{1. Initially, the episode and horizon lengths are equal. 2. Once the agent has learned part of the task, the episode lengths are smaller than the horizon. 3. Reward hacking scenario with 2 active objects where the agent maximizes the return over the horizon length, rather than episode's duration. Dotted line shows desired behaviour, however, the agent extends the episode instead.}
  \label{fig:horizon}
\end{figure}

If the episode and horizon lengths are equal and the agent cannot achieve the goal (i.e. at the beginning of training), the asynchronous experience gathering is equivalent to the synchronous method. However, when the agent starts achieving the goal with higher accuracy, the environment resets are out of sync, therefore the horizon window can span multiple episodes (see Fig. {\ref{fig:horizon}}).

Within the second setup discussed in this section, the agent finds a reward hack during the second phase of the curriculum. A policy is found that can generate the most reward in a window by placing the first box, but not ending the episode with the second box. Instead, it displaces the first box using the second, refusing to finish the episode early such that it circumvents the early negative reward of a fresh episode.

In order to incentivise the agent to finish the episode, we increase the bonus reward for achieving any sub-goal, including the overall environment goal. This, however, requires a balance, since, if this number is set too high, the agent converges on a local optimum, thereby throwing the objects toward the target locations, without actually building a tower.

The second curriculum implementation takes considerably longer to train when compared to the baseline of the staggered reward, introduced in Section \ref{s:ir_results}. Our hypothesis is that an increase in horizon length leads to a drastic change in the expected reward for each state. The returns are approximated using the GAE \cite{schulman2015gae} method, and the value of a state is estimated based on the amount of reward that can be gathered until the end of the episode, starting from that state. Since the agent can now explore more states, due to the increase in maximum timesteps, the expected reward also increases in magnitude. The agent cannot progress on learning the requirements for the new sub-goal until the value function is updated based on the new horizon length. This additional time required is what we believe is slowing down training.

\begin{figure}
    \centering
    \includegraphics[width=.6\textwidth]{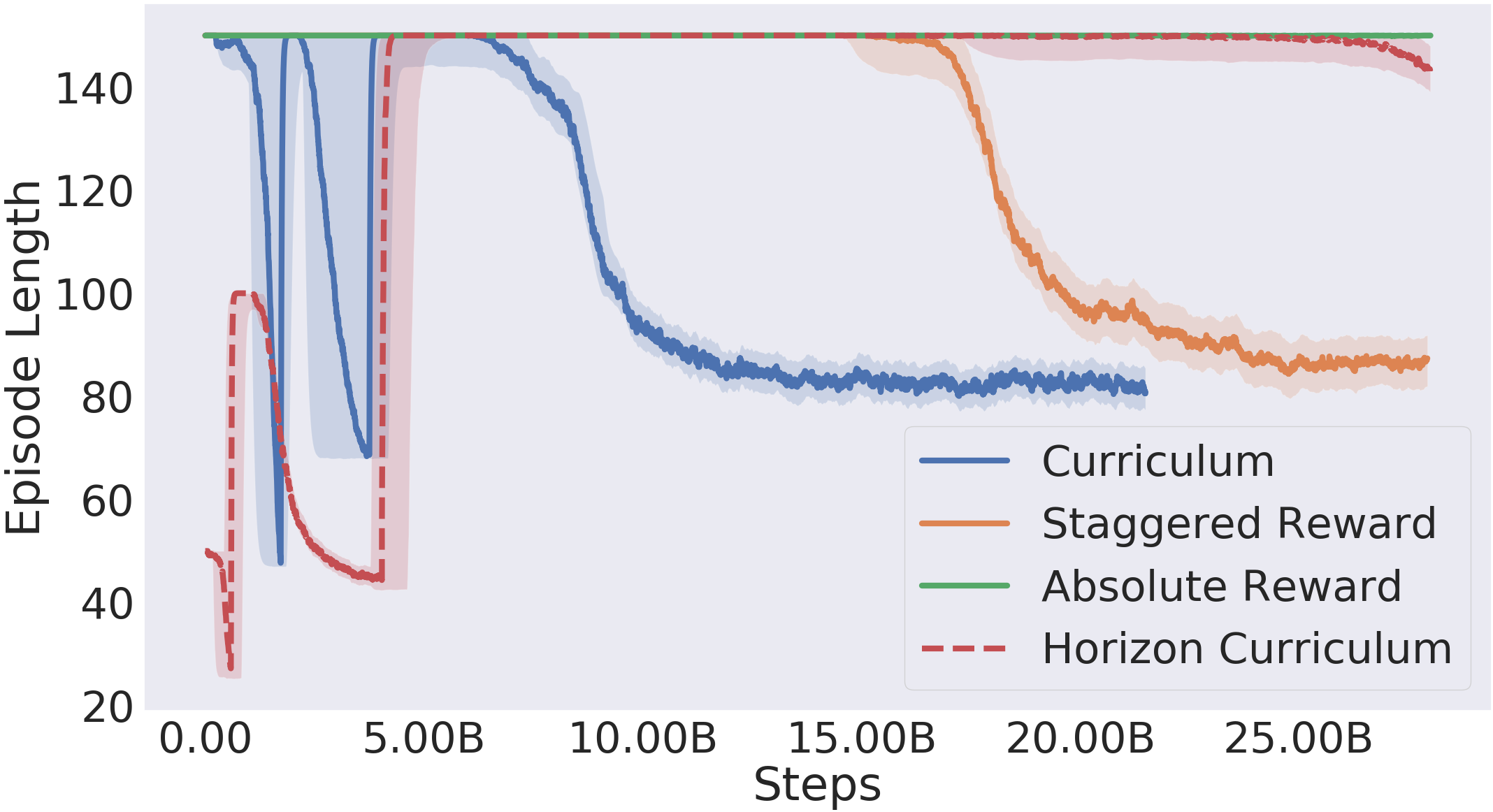}
    \caption{Comparison of curriculum setups against the chosen baselines. Curriculum - Curriculum 1c with H = 150. Staggered and Absolute reward - as described in Table \ref{tab:curriculum1} with H = 150. Horizon Curriculum - second curriculum setup with H = [50, 100, 150]. Curriculum (blue) is the only run achieving 90\% accuracy.}
    \label{fig:all_results}
\end{figure}

To test this, we consider an alternative setup, where $H$ is set to a close-to-optimum value for the last stage of the curriculum directly from the beginning. This means that the early phases of the curriculum will take longer to train due to a non-optimal $H$ value for that stage. However, as shown in Fig. {\ref{fig:all_results}}, this change speeds up training in the long run. Therefore, our initial hypothesis that a curriculum of horizon length speeds up training has been disproved. Table {\ref{tab:horizon-ablation}} encapsulates the ablation study that compares different setups for episode and horizon length over 3 billion steps and with 2 \textit{active} objects.

\begin{table}[h!]
\centering
\caption{Second curriculum ablation results with 2 active objects}
\label{tab:horizon-ablation}
\resizebox{0.68\textwidth}{!}{%
\begin{tabular}{llll}
\hline
\textbf{Max Ep. Length} & \textbf{Horizon Length} & \textbf{Accuracy} & \textbf{Average Ep. Length} \\ \hline
100    & 100    & \textbf{90}\% & \textbf{58} \\
50-100 & 50-100 & 67\% & 62 \\
100    & 10     & 0\% & 100 \\
100    & 50-100 & 71\% & 64 \\ \hline
\end{tabular}%
}
\end{table}

\section{Increasing complexity}
\begin{figure}[!t]
    \centering
    \includegraphics[width=0.45\textwidth]{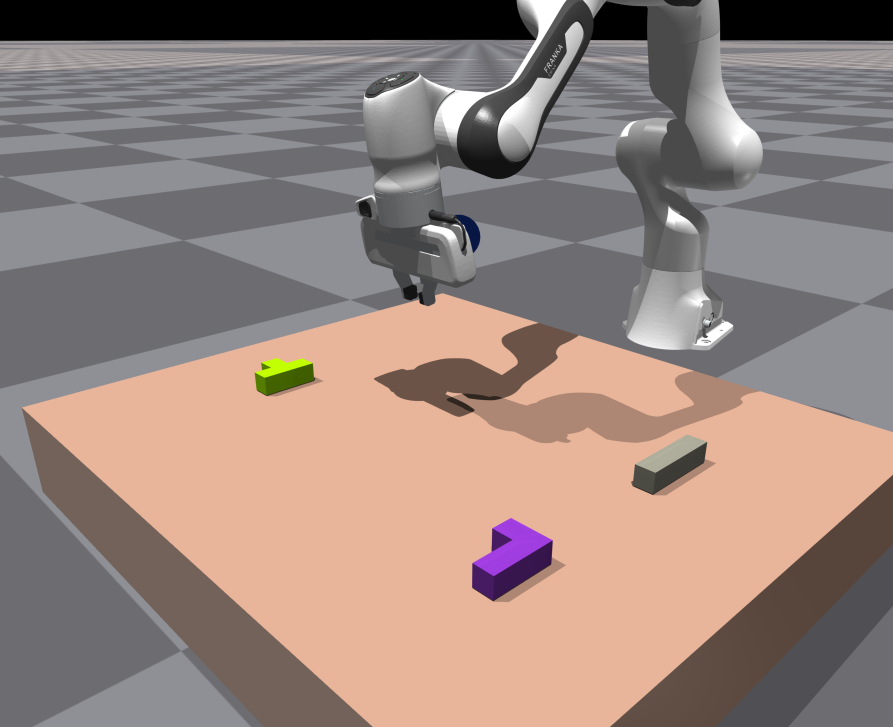}
    \caption{Environment consisting of three pieces from the set of  five tetrominoes (L, T, and I), each being initialised with a fixed orientation. The different shapes increase complexity for the agent.}
    \label{fig:tetris}
\end{figure}

Additionally, we test the first curriculum setup on a more challenging environment (Fig. \ref{fig:tetris}). In this setting, we replace the cubes with three pieces from the set of five tetrominoes (L, T, and I). The increase in complexity comes from the need to learn new different grasping methods associated with each object. To account for this, we increase the network capacity by one layer. In our experiments, the items just need to be stacked in a specified position, although more complexity can be introduced in later experiments, where a certain orientation is required.

We choose the objects to have the same volume as the previously used cubes, however, this led to smaller grasping areas. Initially, the agent fails to learn a good policy within the same time frame. To improve the results, we add a new reward signal $r_{\text{grasp}} = \mathds{1}\{|s_\text{ee} - s_\text{obj}| \leq \epsilon\}$ to assist in guiding the grasping. We find a $\lambda_{\text{grasp}} = 2$ to be most effective.

In a separate test, we increase the complexity further by providing a random initial orientation to each object. To allow the agent to adjust to this change, we concatenate the end-effector's rotation (quaternions) to the observations. The results in Table. \ref{tab:tetris} show that curriculum learning, combined with reward shaping, enables learning even in this complex scenario.

\begin{table}[!ht]
\centering
\caption{Results in the higher complexity environment}
\label{tab:tetris}
\resizebox{0.6\textwidth}{!}{%
\begin{tabular}{llllll}
\hline
Name    & Task        & $\lambda_\text{grasp}$ & Rand. Or. & Acc. & Steps \\ \hline
Cubes   & cubes       & 0                      & no        & 90\% & 16B   \\
Complex & tetrominoes & 2                      & no        & 80\% & 16B   \\
RandOr  & tetrominoes & 2                      & yes       & 59\% & 16B   \\ \hline
\end{tabular}%
}
\end{table}
\section{Limitations}

The trained agents are currently unable to cope with the situation where the tower of stacked objects is knocked down. Instead of trying to reposition the previous objects, the agent continues as though nothing happened. We leave this investigation for future work.

Another limitation of the system presented is that the agent is trained on a maximum of N objects. In order to have N + 1 objects, the training must restart from scratch. In further work, we intend to deal with this scalability issue.

\section{Conclusion}
In this work, we investigated the problem of applying curriculum learning to a goal-oriented environment, such as the stacking of boxes. To accomplish this, we trained an RL agent using a shaped reward function and coupled it with multiple variations of curricula architectures. We found that a simple architecture combined with a curriculum setting can achieve the desired behaviour, whereas vanilla RL fails to complete the task. We provided solutions for a number of reward-hacking scenarios identified and showcased results obtained in a higher-complexity environment. Altogether, these findings establish a strong foundation for future research on CL applied to goal-oriented tasks.

\bibliography{main}

\newpage
\appendix
\onecolumn
\section{APPENDIX} \label{s:appendix}
Code available post review. Hardware used: 2x Nvidia GeForce RTX2080ti.

\setcounter{table}{0}
\renewcommand{\thetable}{A\arabic{table}}

\begin{table}[H]
\centering
\caption{Observation space}
\label{t:obs}
\resizebox{0.7\textwidth}{!}{%
\begin{tabular}{lll}
\hline
\textbf{Observation}  & \textbf{Size} & \textbf{Info}                      \\ \hline
Joint Positions       & 9             & Scaled between -1 and 1            \\
Joint velocities      & 9             &                                    \\
End-effector position & 3             &                                    \\
Object states         & 39            & Position, velocity and orientation \\
Goal position         & 3             &                                    \\
Object id             & 1             & Used for staggered reward          \\ \hline
Total                 & 64            &                                    \\ \hline
\end{tabular}
}
\end{table}

\begin{table}[H]
\caption{Hyperparameters}
\label{t:hyperparams}
\centering
\resizebox{0.4\textwidth}{!}{%
\begin{tabular}{ll}
\hline
\textbf{Hyperparameter} & \textbf{Value}      \\ \hline
Algorithm               & PPO                 \\
Separate Actor/Critic   & False               \\
MLP size                & {[}256, 256, 256{]} \\
Fixed $\sigma$            & False               \\
Activations             & ELU                 \\
Learning rate ($\eta$)    & 5e-4                \\
Learning rate type      & Fixed               \\
Discount factor $\gamma$  & 0.99                \\
GAE $\tau$                & 0.95                \\
KL threshold            & 0.008               \\
Nr environments         & 8192 - 16384               \\
Minibatch size          & 32768               \\
Horizon/Ep length       & 150                 \\ \hline
\end{tabular}
}
\end{table}

\begin{figure}[H]
  \centering
    \includegraphics[width=.53\textwidth]{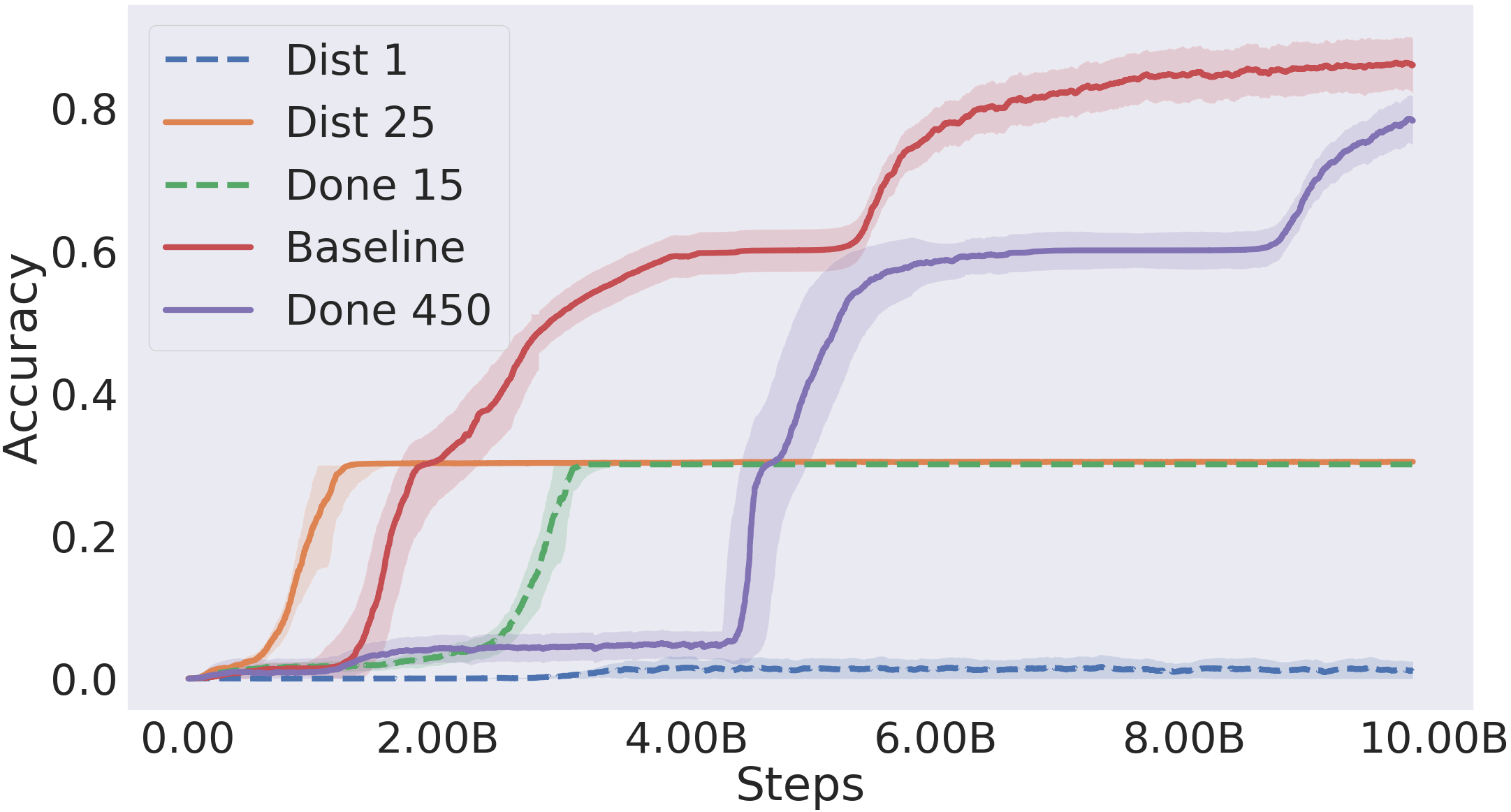}
    \caption{Comparison of various $\lambda$ magnitudes affecting distance-based dense and sparse rewards. 'Done 15' and 'Dist 25' successfully solve stage 1 but fail to progress further. 'Dist 1' fails to find any optimal policy.}
    \label{fig:reward_ablation}
  
\end{figure}%

\subsection*{Reward function ablation} \label{s:rfunction}

We have conducted an ablation study for the values of $\lambda$ used in adjusting our reward function. The values adjusted include the 'Box to Goal' and 'End-effector to Box' distances and the 'Goal achieved' and 'One time bonus per sub-goal' sparse rewards. These are represented in Figure \ref{fig:reward_ablation} as 'Dist' and 'Done', respectively. The baseline consists of $\lambda_\text{dist}=5$ and $\lambda_\text{done}=150$, which represent the values we have used throughout our experiments.

\end{document}